\documentclass[journal]{IEEEtran}

\usepackage{times}
\usepackage{epsfig}
\usepackage{graphicx}
\usepackage{amsmath}
\usepackage{amssymb}
\usepackage{lipsum}
\usepackage[dvipsnames]{xcolor}
\usepackage{multirow}
\usepackage{booktabs}
\usepackage{algorithm}
\usepackage{algorithmic}
\usepackage{amsmath,amssymb}
\usepackage{bm}
\usepackage{amsmath}
\newcommand{\mat}[1]{\bm{#1}}
\newcommand{\ten}[1]{\bm{\mathcal{#1}}}
\usepackage{tabu}
\usepackage{amsthm}
\usepackage{lipsum}
\usepackage{adjustbox}
\theoremstyle{definition}
\newtheorem{definition}{Definition}
\usepackage{wrapfig}
\usepackage{url}
\usepackage{hyperref}

\begin{document}

\title{Lite it fly: An All-Deformable-Butterfly Network}

\author{Rui Lin$^{*,1}$, Jason Chun Lok Li$^{*, 2}$, Jiajun Zhou$^2$, Binxiao Huang$^2$, Jie Ran$^2$ and Ngai Wong$^2$\\ $^1$ Distributed and Parallel Software Lab, Huawei\\ $^2$ Department of Electrical and Electronic Engineering, The University of Hong Kong\\ linrui19@huawei.com, jasonlcl@connect.hku.hk, \{jjzhou, bxhuang, jieran, nwong\}@eee.hku.hk 
\thanks{* Equal contributions}}



\maketitle

\begin{abstract}
Most deep neural networks (DNNs) consist fundamentally of convolutional and/or fully connected layers, wherein the linear transform can be cast as the product between a filter matrix and a data matrix obtained by arranging feature tensors into columns. The lately proposed deformable butterfly (DeBut) decomposes the filter matrix into generalized, butterfly-like factors, thus achieving network compression orthogonal to the traditional ways of pruning or low-rank decomposition. This work reveals an intimate link between DeBut and a systematic hierarchy of depthwise and pointwise convolutions, which explains the empirically good performance of DeBut layers. By developing an automated DeBut chain generator, we show for the first time the viability of homogenizing a DNN into all DeBut layers, thus achieving an extreme sparsity and compression. Various examples and hardware benchmarks verify the advantages of All-DeBut networks. In particular, we show it is possible to compress a PointNet to $< 5\%$ parameters with $< 5\%$ accuracy drop, a record not achievable by other compression schemes.
\end{abstract}


\IEEEpeerreviewmaketitle

\section{Introduction}
\label{sec:intro}


The shifting of AI from cloud to edge or terminal devices with limited computing and storage resources has become another wave of this century and a vital research area. The workhorse operation in DNNs is linear transform, which features various economic variants, e.g., Fourier transform~\cite{bracewell1986fourier}, low-rank~\cite{phan2020stable,yu2017compressing,wang2021kronecker} and sparse matrices~\cite{frankle2018lottery,frankle2020linear,wu2019compressing}. In order to obtain compact DNNs with fewer parameters for edge AI implementation, various specialized structured linear transforms are developed to substitute the fully-connected (FC) and/or convolution (CONV) layers. Under the scope of structured \textbf{sparse} linear transforms, Fastfood Transform~\cite{le2013fastfood} is a representative one that belongs to the category of kernel methods. It approximates a dense Gaussian random matrix by computing the product between Hadamard and diagonal Gaussian matrices, thus reducing the number of parameters in representing the FC layers. However, its diagonal Gaussian matrices are not learnable so that Adaptive Fastfood~\cite{yang2015deep} is proposed which updates the elements of the diagonal Gaussian matrices through training. Lately, a special kind of block-diagonal matrices with a recursive structure called Butterfly is proposed~\cite{dao2019learning}. An extension named Kaleidoscope matrix~\cite{dao2020kaleidoscope} then follows, which multiplies the original Butterfly series with its transposed correspondent when approximating the weights matrix of an FC layer.


\begin{figure}[t]
\centering
\includegraphics[scale=1]{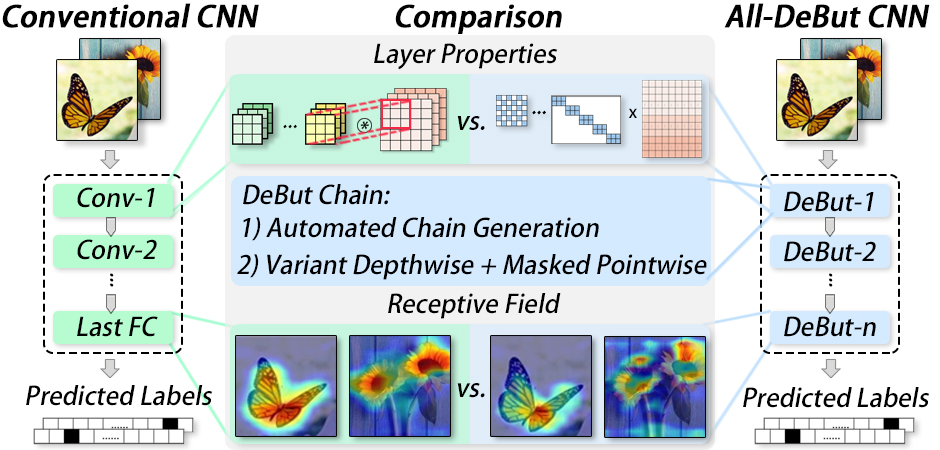}
\caption{In this work, we obtain the homogeneous Deformable Butterfly (DeBut) network, whose layers are \textbf{\textit{all}} replaced by a series of highly structured sparse matrices designed \textbf{\textit{automatically}}. It is worth noting that we further \textbf{\textit{decipher the relations}} between DeBut and the widely utilized depthwise separable convolution.
}
\label{fig:idea_sum}
\vspace{-5mm}
\end{figure}

Nonetheless, the above schemes are only applied to replace the FC layer using a square weight matrix whose size is constrained to be powers-of-two (PoT). Otherwise, the weight matrix will be padded by zeros, and the redundant part of the output is discarded, leading to a waste of compute and storage. To overcome these,~\cite{lin2021deformable} generalizes the standard Butterfly and defines a new type of structured matrices called Deformable Butterfly (DeBut) (cf. Definition~\ref{def:DeBut_notation}) with flexible shapes. Furthermore, under the {\tt im2col} settings (cf. Definition~\ref{def:im2col}), both FC and CONV layers can be replaced by the DeBut chains (cf. Definitions~\ref{def:DeBut_chain_mono}\&\ref{def:DeBut_chain_bulging}) without the PoT limitation. 

On the other hand, the origin of CNNs can be dated back to the 1980s~\cite{fukushima1982neocognitron}, which gained popularity after AlexNet~\cite{krizhevsky2012imagenet} achieved remarkable performance on ILSVRC-2012 ImageNet dataset. These CNNs, however, are based on heterogeneous modules that alternate between convolutional and pooling layers, followed by FC layers. It was not until~\cite{springenberg2014striving} that the concept of a homogeneous network consisting of solely CNN layers was explored. Inspired by this All-CNN network \cite{springenberg2014striving}, in this work, we explore the potential of All-DeBut networks wherein all layers are substituted by their DeBut counterparts. While one or a few DeBut layers can be handcrafted, replacing all layers with their DeBut formats quickly become intractable. Therefore, the automated chain generation scheme proposed in this paper is crucial for practical reasons, which largely eliminates the need for labor-intensive design for each DeBut chain. A major merit of an All-DeBut network arises from its extreme sparsity and the potential benefits from an implementation perspective. In this regard, we establish a link between an all-DeBut layer and a generalized depthwise-separable CNN layer, followed by a hardware deployment on FPGA to confirm its hardware friendliness. Our key contributions are:
\begin{itemize}
\item Demonstrating the feasibility of All-DeBut networks for the \textbf{first} time, whose layers are \textbf{all} cast into DeBut chains. 
\item An \textbf{automated} chain generation scheme that suits most CONV and FC layers and significantly reduces the labor in handcrafting, making possible the design of an All-DeBut network.
\item Further \textbf{analytical} results of DeBut to explain its empirically excellent performance. In particular, we decipher the relation between a DeBut chain and a non-trivial depthwise-separable-like convolution. 
\item FPGA  \textbf{deployment} to verify the DeBut's efficacy on resource-constrained devices.
\end{itemize} 
\vspace{-3mm}

\section{Preliminary}
\label{sec:preliminary}
We use the 4-D tensor $\ten{K} \in \mathbb{R}^{k \times k \times C_i \times C_o}$ to denote the weights of a CONV layer, where $k$ represents the spatial size of the kernel window, $C_i$ and $C_o$ are the numbers of input and output channels, respectively. The 3-D tensor $\ten{X} \in \mathbb{R}^{H_i \times W_i \times C_i}$ denotes the corresponding input of the selected layer, where $H_i$ and $W_i$ are the height and width of the input, and $C_i$ is the number of channels. It is worth noting that an FC layer can be regarded as a CONV layer by setting the kernel size equal to the spatial size of the input. For brevity and without loss of generality, we no longer specifically distinguish between the CONV and FC layers in the sequel. We first summarize the important concepts in DeBut in the form of definitions~\cite{lin2021deformable}.

\begin{figure*}[t]
\centering
\includegraphics[scale=0.82]{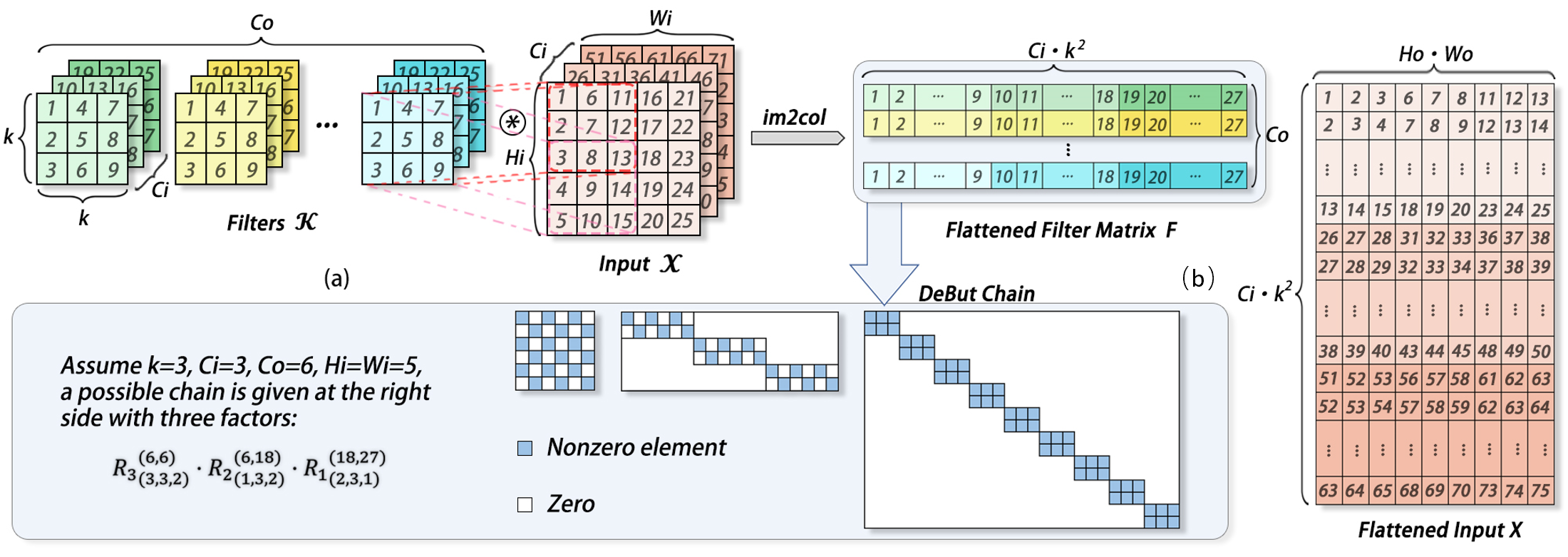}
\vspace{-2mm}
\caption{\textbf{(a)} Standard convolution. \textbf{(b)} Flattened filter and input matrices via the $\tt{im2col}$ operation. A possible DeBut chain for the filter matrix $\mathbf{F}$ is given.}
\label{fig:connection}
\end{figure*}

\begin{definition}
\label{def:im2col}
\textbf{\textit{(im2col operation)}} The {\tt im2col} operation flattens out the feature map entries in a window followed by stacking them as columns in a matrix (cf. Fig.~\ref{fig:connection}).
\end{definition}

\begin{definition}
\label{def:DeBut_notation}
(\textbf{\textit{DeBut matrices}}) A DeBut matrix comprises block matrices along its main diagonal, denoted $R^{(p,q)}_{(r,s,t)} \in \mathbb{R}^{p\times q}$, where $(r,s,t)$ stands for $r \times s$ blocks of $t \times t$ diagonal matrices. The subscript and superscript satisfy $\frac{p}{r t} = \frac{q}{s t}$.
\end{definition}

\begin{definition}
\label{def:DeBut_chain_mono}
(\textbf{{\textit{Monotonic chain}}}) A monotonic DeBut chain is formed by a series of DeBut factors ${R_m}^{(p_m, q_m)}_{(r_m,s_m,t_m)} \times  \cdots \times {R_1}^{(p_1, q_1)}_{(r_1,s_1,t_1)}$ meeting the constraints:%
\begin{subequations}
\label{eq:mono_pro}
\begin{equation}
\label{eq:condition_1}
q_{i+1} = p_i \text{ for } i=1,2, \cdots,m-1
\end{equation}
\begin{equation}
\label{eq:t_rt_1}
t_1 = 1 \text{,\,\,} t_{i+1} = r_i  t_i \text{\,\, and \,\,} \frac{p_m}{r_m  t_m} = \frac{q_m}{s_m t_m} = 1
\end{equation}
\begin{equation}
\begin{aligned}
\forall\, i= 1,2, \cdots,&m-1 \\ \text{ s.t. } p_i \le q_i \text{~if\, $p_m \le q_1$} &\text{ , or\, } p_i\ge q_i \text{~if\, $p_m \ge q_1$}
\end{aligned}
\end{equation}
\end{subequations}
\end{definition}

\begin{definition}
\label{def:DeBut_chain_bulging}
(\textbf{\textit{Bulging chain}}) A bulging DeBut chain is formed by a series of DeBut factors ${R_m}^{(p_m, q_m)}_{(r_m,s_m,t_m)} \times  \cdots \times {R_1}^{(p_1, q_1)}_{(r_1,s_1,t_1)}$ meeting the constraints Eqs.~(\ref{eq:condition_1}) and (\ref{eq:t_rt_1}), plus:%
\begin{equation}
\label{eq:bulging_pro}
\exists\, i= 1,2, \cdots,m-1 \text{ s.t. } p_i > q_i
\end{equation}
\end{definition} 
To avoid the for-loops when doing convolution in the conventional way depicted in Fig.~\ref{fig:connection}(a), the {\tt im2col} operation is applied on both the filters $\ten{K}$ and the input $\ten{X}$, thus resorting to the more efficient GEMM approach (cf. Fig.~\ref{fig:connection}(b)). We employ $\mat{F}\in \mathbb{R}^{C_o \times C_i  k^2}$ and $\mat{X}\in \mathbb{R}^{C_i   k^2 \times H_o   W_o}$ to denote the flattened filter and input matrices, respectively, where ($H_o$, $W_o$) is the spatial size of the output. By adopting the notation $R^{(p,q)}_{(r,s,t)}$, the product of a sequence of DeBut factors can be represented efficiently. Based on the properties formulated in Eqs.~(\ref{eq:mono_pro}) and~(\ref{eq:bulging_pro}), a given sequence of DeBut factors can be categorized as a monotonic or bulging chain, which will be used to substitute the flattened filter matrix $F$ with structured \textbf{learnable} elements. The lower left block in Fig.~\ref{fig:connection}(b) exemplifies using a monotonic chain with three DeBut factors to approximate a given $\mat{F}\in \mathbb{R}^{6\times 27}$. By replacing a CONV/FC layer with DeBut, the complexity is reduced from $\ten{O}(c_o k^2c_i H_oW_o)$ to $\ten{O}(\max_{i\in\{1,...,N_{factor}\}} p_is_iH_oW_o)$, where $N_{factor}$ is the number of DeBut factors, and $\max_{i \in \{1,...,N_{factor}\}}p_is_i$ denotes the maximum number of nonzeros in a single DeBut factor within the DeBut chain.

\section{All-DeBut Networks}
\label{sec:method}
Here an automated DeBut chain generation scheme that fits most CONV layers is proposed, which eliminates the labor of handcrafting differently sized DeBut chains. Additionally, we provide new exposition and analytical explanation of the benefits brought by the DeBut layers, which largely explain the empirically remarkable performance of DeBut substitutes for linear transform in DNNs. Given a CONV layer, it is impractical to enumerate all possible DeBut chains, while manually designing chains that meet different requirements is time-consuming. To overcome this, we propose an automated chain generation scheme considering typical CONV layer properties. By observing the structures of popular CNNs such as AlexNet~\cite{krizhevsky2012imagenet}, VGGNet~\cite{simonyan2014very}, ResNet~\cite{he2016deep}, etc., we summarize the following properties of common CONV layers:
\begin{itemize}
\item Square kernel windows with spatial size $k=1$, $k=3$, $k=5$, or $k=7$ are the most popular.
\item The number of input channels $C_i$ and output channels $C_o$ of each layer will not change sharply, and there are three widely adopted conventions: $C_o = C_i$, $C_o = \frac{1}{2} C_i$ or $C_o = 2 C_i$.
\item The numbers of input channels $C_i$ and output channels $C_o$ are typically PoT.
\end{itemize}
We focus on generating chains automatically for CONV layers meeting these properties, subject to various compression ratios. A workaround for \textit{arbitrarily} sized chains is to use zero padding to fill up the matrix to meet the conditions.

As introduced in~\cite{lin2021deformable}, a DeBut chain can be categorized as a monotonic or a bulging chain. Besides, the chains in the same category can exhibit different compression ratios due to different shrinking speeds. To quantify the chain design requirement and to facilitate coding, we define three hyperparameters, namely, the shrinking level $N \in \mathbb{Z}^+$, the bulging rate $\alpha \in \mathbb{R}$ $(\alpha > 1)$ needed for bulging chain design, and a collection of optional shapes of the diagonal blocks $\ten{S} = \{(r,s)|r,s\in \mathbb{Z}^+ \}$. The shrinking level $N$ is related to the number of DeBut factors. The larger the value of $N$, the more DeBut factors and the slower the shrinkage. The bulging rate can decide the chain's feature representation capacity which is proportional to $\alpha$. Motivated by the limited search space in Neural Architecture Search (NAS), which improves the searching efficiency significantly, we use the predefined $\ten{S}$ to constrain the shape of the diagonal blocks. Every tuple of $(r,s)$ in $\ten{S}$ describes the partition of the diagonal blocks. Since $\ten{S}$ contains only a limited number of optional block shapes, it largely simplifies the complexity of the chain generation.

\subsection{Automated Chain Generation}
\label{subsec:autogen}
\begin{table*}[t]
\scriptsize
\centering
\caption{Four successive stages to determine the superscripts of the factors when designing chains under different requirements. The number of factors generated in each stage is provided, where $k$ is the kernel size, $N$ is the shrinking level, and $\alpha$ is the bulging rate. It is worth noting that the four successive stages do not determine the inner structure of the factors, namely, $(r, s)$, which is dependent on the pre-defined searching pool \ten{S}.}
\label{tab:S_sup_rule}
\setlength{\tabcolsep}{0.8mm}{
\begin{tabular}{l|l|l|l|l|l}
\toprule
\multirow{2}{*}{Type}  & \multirow{2}{*}{Factor Shape} & \multicolumn{3}{c|}{\# Factors} & \multirow{2}{*}{Stage Description} \\
\cline{3-5}
~ & ~ & \multicolumn{1}{l|}{C\_o = C\_i} & \multicolumn{1}{l|}{C\_o = C\_i / 2} & \multicolumn{1}{l|}{C\_o = 2 C\_i} & ~  \\ 
\midrule
\multirow{4}{*}{Mono.} & $[2^{k+\log_2^{C_i}}, W]$ & $1$ & $1$ & $1$ & Stage 1: Initialize the first factor. \\
~ & $[2^{k+\log_2^{C_i}}, 2^{k+\log_2^{C_i}}]$ & $\max(0, N-5)$ & $\max(0, N-5)$ & $\max(0, N-3)$ & Stage 2: $N$ will determine \#Factors with the same shape.\\
~ & $[2^{k+\log_2^{C_i}-j-1}, 2^{k+\log_2^{C_i}-j}]$ & 2 $(j \in [0, 1])$ & 3 $(j \in [0, 1, 2])$ & 1 $(j=0)$ & Stage 3: Generate the monotonous factors.  \\
~ & $[H, 2  H]$ & $1$ & $1$ & $1$ & Stage 4: Generate the final factor. \\
\midrule
\multirow{4}{*}{Bulg.} & $[2^{k+\log_2^{C_i}}  \alpha, W]$ \& $[2^{k+\log_2^{C_i}}, 2^{k+\log_2^{C_i}}  \alpha]$  & $2$ & $2$ & $2$  & Stage 1: Initialize the first factor.\\
~ & $[2^{k+\log_2^{C_i}}, 2^{k+\log_2^{C_i}}]$ & $\max(0, N-6)$ & $\max(0, N-6)$ & $\max(0, N-4)$ & Stage 2: $N$ will determine \#Factors with the same shape.\\
~ & $[2^{k+\log_2^{C_i}-j-1}, 2^{k+\log_2^{C_i}-j}]$ & $2$ $(j \in [0, 1])$ & $3$ $(j \in [0, 1, 2])$ & $1$ $(j=0)$ & Stage 3: Generate the bulging factors.\\
~ & $[H, 2H]$ & $1$ & $1$ & $1$ & Stage 4: Generate the final factor.\\
\bottomrule
\end{tabular}}
\end{table*}

Based on the observation and the hyperparameters, we propose the automated chain generation scheme, which records the superscripts $S_{sup}$ and subscripts $S_{sub}$ of the factors. It contains three core parts: 1) check the size of kernel tensor $\ten{K}$; 2) make the rule to get $S_{sup}$; and 3) use $\ten{S}$ to determine the shape of diagonal blocks. Algorithm~\ref{alg:auto_chain_sum} is the simplified version of pseudo-code for easier understanding, and the detailed version and its practical implementation are available in \href{https://github.com/jasonli0707/auto_debut_chain_generator}{our GitHub}, which has a time complexity of $O(n)$ with $n$ being the number of DeBut factors. For example, designing a 7-factor chain for a matrix of size $[256, 2304]$ only takes $1.7$ms. The following is an explanation of the three key components. In succinct terms, the proposed scheme executes a particular factorization obeying the rules in Definitions~\ref{def:DeBut_chain_mono} \& ~\ref{def:DeBut_chain_bulging}.

\textbf{Check the size of the kernel tensor $\ten{K}$.} To instantiate a reasonable search space for DeBut chain generation, a PoT constraint is imposed on the numbers of input and output channels, viz. $C_i$ and $C_o$, of the selected layer. In case of exceptions, we roundup $C_i$ and $C_o$ to their nearest PoT numbers via a ceiling function ${\rm{ceil}(\circ)}$, namely, $\hat{C}_i = 2^{\rm{ceil}(\log_2^{C_i})}$ and $\hat{C}_o = 2^{\rm{ceil}(\log_2^{C_o})}$.

\textbf{Formulate the rules to design $\mathbf{S_{sup}}$.}  In our scheme, the determination of the superscripts of the factors when designing a chain can be divided into four successive stages. For clarity, we systematically list the four stages and their design rules in Table~\ref{tab:S_sup_rule} under different conditions. We remark that the superscripts is related to the kernel size $k$, and the shrinking level $N$ is utilized to adjust the number of factors. By adjusting the value of $\alpha$, bulging chains with distinct representation powers can be generated. 

\textbf{Determine the partition of diagonal blocks by $\ten{S}$.} After formulating the rules to obtain $S_{sup}$ (cf. Table~\ref{tab:S_sup_rule}), the relation between the height and width of each DeBut factor is fixed. Therefore, we can design a set of optional shapes for the diagonal blocks. For example, we can set $(r,s)$ for a factor of shape $[2^{k+\log_2^{C_i}-j-1}, 2^{k+\log_2^{C_i}-j}]$ to be $[2, 4]$, $[4, 8]$, or $[16, 32]$, etc. Similarly, we can prepare $(r,s)$ for factors in other stages. In the chain generation, $\ten{S}$ works as a pool to provide multiple settings of the diagonal blocks.

\begin{algorithm}
    \footnotesize
    \renewcommand{\algorithmicrequire}{\textbf{Input:}}
    \renewcommand{\algorithmicensure}{\textbf{Output:}}
    \caption{Automated Chain Generator (Simplified Version)}
    \label{alg:auto_chain_sum}
    \begin{algorithmic}[1]
    \REQUIRE 4-D kernel tensor $\ten{K}$ of the selected CONV layer; shrinking level $N$; type of the DeBut chain $\{ \rm{'mono'}, \rm{'bulging'} \}$; bulging rate $\alpha$ for bulging chain design; and a collection of optional shapes of the diagonal blocks $\ten{S}$.
    \ENSURE The compression ratio $\eta$; the superscripts $S_{sup}$; and the corresponding subscripts $S_{sub}$.
    \STATE $C_o, C_i, k, k = \ten{K}$.shape $\leftarrow$ Get the shape of the kernel.
    \STATE $C_o = 2^{ceil(\log_2^{C_o})}$, $C_i = 2^{ceil(\log_2^{C_i})}$ $\leftarrow$ If the number of output/input channels is not the PoT, we force it to be the nearest PoT.
    \IF {type == 'mono'}
    \STATE $S_{sup}$.$\rm{append}$([$\rm{int}$($2^{k+\rm{log}_2^{C_i}}$), W]), $S_{sub}$.$\rm{append}$([$2^k,k^2,1$]) $\leftarrow$ \textbf{Stage 1}: Determine the structure of the rightmost DeBut factor.
    \STATE $S_{sup}.\rm{append}([2^{k+\rm{log}_2^{C_i}}, 2^{k+\rm{log}_2^{C_i}}])$, $S_{sub}.\rm{append}([r,s,t])$ where $(r, s) \in \ten{S}$ $\leftarrow$ \textbf{Stage 2}: Generate \#Factors (determine by $N$) with the same shape, and search $(r, s)$ in the pre-defined \ten{S}.
    \STATE $S_{sup}.\rm{append}([2^{k+\rm{log}_2^{C_i}}, 2^{k+\rm{log}_2^{C_i}-j-1}]$, $S_{sub}.\rm{append}([r,s,t])$ where $(r, s) \in \ten{S}$ $\leftarrow$ \textbf{Stage 3}: Generate the monotonous factors.
    \STATE $S_{sup}.{\rm{append}}([{\rm{int}}(H), {\rm{int}}(2   H)])$, $S_{sub}.\rm{append}([r,s,t])$ where $(r, s) \in \ten{S}$ $\leftarrow$ \textbf{Stage 4}: Generate the final factor.
    \ENDIF
    \IF {type == 'bulging'}
    \STATE Follow the pipeline described above but use the pre-defined stages for bulging chains in Table~\ref{tab:S_sup_rule}.
    \ENDIF
    \STATE $\eta$ $\leftarrow$ Count the number of nonzeros $N_{non}$ and compute the compression ratio.
    \STATE \textbf{return} $\eta$, $S_{sup}$, $S_{sub}$ $\leftarrow$ Return the compression ratio and the automatically designed DeBut chain.
    \end{algorithmic}
\end{algorithm}

\subsection{Connection to Depthwise Separable Convolution}
\label{subsec:conv}
\begin{figure*}[t]
\centering
\includegraphics[scale=0.85]{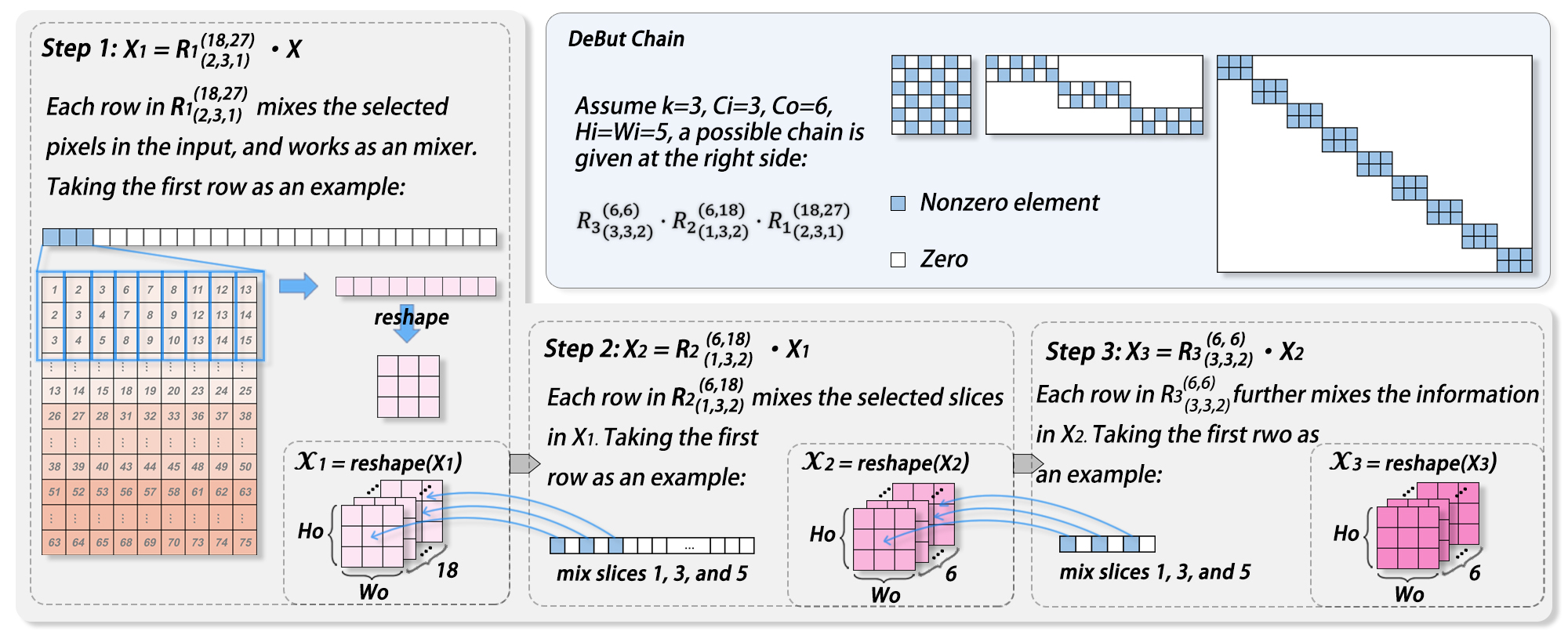}
\caption{Computation flow along the DeBut chain, connecting the Debut layer to the depthwise separable convolution operations. With a given chain, \textbf{Step 1} shows how the first row in {\scriptsize ${R_1}_{(2,3,1)}^{(18,27)}$} works as a mixer to merge the information in the first channel, and the remaining rows work similarly. By reshaping the merged information vector obtained by each row respectively, $\mat{X}_1$ can be regarded as a 3-D tensor $\ten{X}_1$. In $\textbf{Step 2}$, the example illustrates that each row in {\scriptsize ${R_2}_{(1,3,2)}^{(6,18)}$} works as a pointwise convolution while the weights assigned for some slices are zeroed. Likewise, ${\textbf{Step 3}}$ further mixes the information, which allows each pixel in an output channel to contain the information from all pixels in the input feature tensor.}
\label{fig:connection_part2}
\vspace{-3mm}
\end{figure*}

MobileNet~\cite{howard2017mobilenets} and EfficientNet~\cite{tan2019efficientnet} are two favored lightweight models with impressive performance, wherein the depthwise separable convolution is celebrated for its representation capability with much fewer parameters than standard {\tt conv2d} convolution. To decode why DeBut can replace CONV layers with dramatically fewer parameters while maintaining a decent accuracy, we explore the link of DeBut to a special, modified form of depthwise separable convolution which comprises depthwise and pointwise convolutions. Specifically, the conventional convolution is depicted in Fig.~\ref{fig:connection}(a), every kernel (filter) in the CONV layer will slide along the spatial axes (viz. $H$ and $W$ axes) of the input, doing convolution with the corresponding input marked in the kernel window across the $C_i$ axis. Without loss of generality, applying column-major ${\rm{im2col}}$ operation on the filters and input, the flattened filter matrix $\mat{F}\in \mathbb{R}^{C_o \times C_i k^2}$ and input $\mat{X} \in  \mathbb{R}^{C_i k^2 \times H_o  W_o}$ are obtained. The matrix multiplication shown in Fig.~\ref{fig:connection}(b) is equivalent to the convolution displayed in Fig.~\ref{fig:connection}(a). For ease of illustrating the aforementioned convolution operations, a simple DeBut chain with three factors is exemplified in Fig.~\ref{fig:connection}(b), which is under the assumption $k=3$, $C_i=3$, and $C_o=6$. Fig.~\ref{fig:connection_part2} visualizes the calculation between the three-factor chain and the flattened input $\mat{X}$.

\textbf{DeBut vs Depthwise Convolution.} The first step shown in Fig.~\ref{fig:connection_part2} displays the multiplication between the rightmost DeBut factor ${R_1}_{(2,3,1)}^{(6,18)}$ with $\mat{X}$, taking the first row as an example. It is seen that the nonzero elements in each row can mix the information selected in every column of $\mat{X}$. There are three cases of the relations between $s_1$ (i.e., the number of nonzeros in each row of the rightmost DeBut factor) and the size of the kernel window $k$: 1) $s_1 < k^2$, 2) $s_1 = k^2$, and 3) $ s_1 > k^2$. In all cases, it is noted that if $r>1$, it means more than one kernel is assigned to mix the elements at the same position. 

\textit{Case 1 (${s_1 < k^2}$):} Each row of the rightmost factor mixes \textbf{part} of the pixels in the kernel window when it slides along the $H$ and $W$ axes of a selected channel, which can be regarded as a \textbf{sub-sampling} version of depthwise convolution (cf. Step 1 in Fig.~\ref{fig:connection_part2}). Besides, some rows in the DeBut factor have the chance to merge information from two different channels when $k^2$ is not an integer multiple of $s_1$. We remark that sub-sampling the pixels in the kernel window will not degrade the representation capability of the rightmost DeBut factor, since \textbf{all} pixels will be sampled due to the \textbf{complementary} positions of the nonzeros in rows from different diagonal blocks. 

\textit{Case 2 (${s_1 = k^2}$):} Each row in the rightmost DeBut factor is identical to depthwise convolution in a fixed channel. When $r>1$, it means more than one kernel is assigned to a channel, thus the rightmost DeBut factor is expected to generate intermediate output with richer information.

\textit{Case 3 (${s_1 > k^2}$):} It can be treated as a counterpart of the first case, being an \textbf{up-sampling} depthwise convolution. Therefore, the rightmost DeBut factor belonging to this case also warrants informative feature maps as well.

\textbf{DeBut vs Pointwise Convolution.} For better understanding, we reshape $\mat{X}_1$ obtained in Step 1 into a 3-D tensor, and denote it as $\ten{X}_1$. Each slice in $\ten{X}_1$ is the merged information obtained by a distinct row in the rightmost DeBut factor. The second step in Fig.~\ref{fig:connection_part2} takes the first row in the middle DeBut factor as an example to show the connection between DeBut and pointwise convolution. It is seen that the nonzeros in each row are the weights assigned to the corresponding slices in $\ten{X}_1$, which can be regarded as a \textbf{masked} version pointwise convolution. Notice that no information will be lost thanks to the complementary positions of the nonzeros. The final step in Fig.~\ref{fig:connection_part2} can be regarded as a masked pointwise convolution, which further merges the information based on the reshaped $\mat{X}_2$, filling each entry in the final output with information from every entry in the input tensor. Consequently, given a DeBut chain with $m$ factors, the multiplication with the flattened input can be understood as a \textbf{variant} of depthwise convolution followed by $(m-1)$ \textbf{masked} pointwise convolution, whose final output preserves all information from the input tensor.



\subsection{Feature Distance and Compression Flexibility}
\label{subsec:grad_cam}
Here we focus on the distance between the feature maps obtained by the original CNN and its All-DeBut counterpart, which demonstrates how DeBut affects the performance of the compact model and shows the compression strength of the All-DeBut network. We omit nonlinear operations like activation, max pooling, and batch normalization for brevity when formulating the distance. From the {\tt im2col} view, the feature map of the original CNN after the forward pass when given the input $\ten{X}$ can be described by:
\begin{equation}
\footnotesize
\label{eq:feature_original}
f_1(\ten{X}) = \left(\prod_{l=1}^L \mat{F}_l\right) \mat{X},
\end{equation}
where $L$ is the number of layers of the CNN, $\mat{F}_l$ the flattened kernel matrix of layer $l$, and $\mat{X}$ the flattened input. For its All-DeBut correspondence, the final feature map is: 
\begin{equation}
\footnotesize{
\label{eq:feature_all_debut}
f_2(\ten{X}) = \left(\prod_{l=1}^{L}\prod_{j=1}^{m_l} {R_j^l}^{(p_j^l,q_j^l)}_{(r_j^l, s_j^l, t_j^l)}\right) \mat{X}},%
\end{equation}
where $m_l$ is the number of factors of the DeBut chain that substitutes layer $l$, and ${R_j^l}^{(p_j^l,q_j^l)}_{(r_j^l, s_j^l, t_j^l)}$ $(j=1,2,\cdots,m_l)$ represent the specific factors.

The Kullback-Leibler (KL) divergence is employed to evaluate the difference between $f_1(\ten{X})$ and $f_2(\ten{X})$. The smaller the distance, the more likely the compressed network will obtain the same inference results. This KL distance reads:
\begin{equation}
\footnotesize
\label{eq:feature_kld}
\begin{aligned}
& D_{KL}(f_1(\ten{X})||f_2(\ten{X})) \\
=& \left( \prod_{l=1}^L \mat{F}_l \right) \mat{X} \log \left(\frac{(\prod_{l=1}^L \mat{F}_l) \mat{X}}{(\prod_{l=1}^{L}\prod_{j=1}^{m_l} {R_j^l}^{(p_j^l,q_j^l)}_{(r_j^l, s_j^l, t_j^l)}) \mat{X}} \right) \\
\end{aligned}
\end{equation}
It is seen that the distance is jointly determined by the approximation of all substituted layers. It is worth noting that Alternating Least Squares (ALS) is utilized to minimize the distance in~\cite{lin2021deformable} at the initialization. However, Eq.~(\ref{eq:feature_kld}) can be considered during the training. In this paper, we explore the use of knowledge distillation as a generic training framework for All-DeBut networks, with the goal of closing the gap between the extremely sparse DeBut student and the teacher of its choice. Specifically, we select Contrastive Representation Distillation (CRD) \cite{tian2019crd}, which has shown empirically good results through extensive experiments as compared to other distillation approaches. According to Eq.~(\ref{eq:feature_kld}), by assigning DeBut chains to different layers relying on their properties (e.g., position, and kernel size), various All-DeBut networks meeting different compression requirements can be obtained. 

\section{Experiments}
\label{sec:exp}
To demonstrate the feasibility of the homogeneous All-DeBut network, extensive experiments are conducted with ModelNet40~\cite{wu2015modelnet}, CIFAR-100~\cite{krizhevsky2009cifar100} and ImageNet~\cite{deng2009imagenet} as datasets. First, we verify the validity of the automated chain generation scheme using VGG~\cite{simonyan2014very}. Then we employ the popular PointNet~\cite{qi2017pointnet} for ModelNet40, and choose ResNet~\cite{he2016deep} for CIFAR-100 and ImageNet. Since DeBut is orthogonal to popular compression approaches like quantization and pruning, we focus on comparing All-DeBut with other homogeneous networks obtained by using SVD, Adaptive Fastfood~\cite{yang2015deep}, and Standard Butterfly~\cite{dao2019learning} to substitute the layers. We emphasize that homogeneous neural networks can be classified into two categories: 1) adjustable model size, and 2) fixed model size. All-DeBut and All-SVD are examples of the former class, while All-Butterfly and All-Adaptive Fastfood belong to the second category. Additionally, FPGA benchmarking results of DeBut are also provided to confirm its capability in hardware acceleration.


\begin{table}[ht]
\centering
\scriptsize
\caption{Results on CIFAR-100 with VGG16-BN, using auto-generated DeBut chains of different shrinking levels $N$. Model-wise compression is abbreviated as ``MC''.}
\label{tab:exp_auto_chain}
\setlength{\tabcolsep}{1.8mm}{
\begin{tabular}{lccccc}
\toprule
Model & $N$ & MC (\%) & \#Params & Top-1 Acc. (\%) & Top-5 Acc. (\%)\\
\midrule
VGG16-BN & -- & -- & $14.77$M & $75.09$ & $92.74$\\
All-DeBut & $3$ & $94.40$ &  $0.83$M & $70.27$ & $92.04$\\
All-DeBut & $5$ & $94.74$ & $0.78$M & $68.70$ & $91.45$\\
All-DeBut & $7$ & $95.90$ & $0.61$M & $61.70$ & $88.17$\\
\bottomrule
\end{tabular}}
\vspace{-4mm}
\end{table}

\subsection{Automated Chain Generation}
\label{subsec:autochain}
First, we demonstrate the feasibility of the automated chain generation by applying it to VGG~\cite{simonyan2014very}, and evaluate its performance on CIFAR-100~\cite{krizhevsky2009cifar100} dataset. Table~\ref{tab:exp_auto_chain} summarizes the performance of the All-DeBut networks obtained by different sets of DeBut chains. Here, we fix the chain type to monotonic and set different shrinking levels $N$ to generate chains automatically.

According to Table~\ref{tab:exp_auto_chain}, it is seen that the larger the shrinking level $N$, the higher the compression. By employing the automatically generated chains, the All-DeBut network can achieve impressive model-wise compression ratios with acceptable accuracy loss. Subsequently, the automated chain generation scheme offers us a way to quickly prototype CNNs with DeBut layers as well as flexibility to vary the model size, which is an advantage over other fixed-size structured linear transforms such as Butterfly~\cite{dao2019learning} and Adaptive Fastfood~\cite{yang2015deep}.

\subsection{All-DeBut PointNet on ModelNet40}
\label{subsec:pointnet_ModelNet40}
The advancement in 3D sensory devices, such as LIDAR, has made 3D data increasingly accessible and offered rich geometric information for robotics, autonomous driving, and many more. The \textbf{ModelNet40} dataset is a widely adopted benchmark for 3D point cloud classification. It comprises of 12311 CAD models separated into 9843 training and 2468 test samples from 40 manufactured object categories. In this experiment, we employ the popular PointNet as our backbone.


\begin{table}[ht]
\caption{Results on ModelNet40 of homogeneous networks obtained via different linear transforms, with PointNet backbone.}
\label{tab:exp_1}
\scriptsize
\centering
\setlength{\tabcolsep}{1.6mm}{
\begin{tabular}{lcccc}
\toprule
Method & MC (\%) & \#Params & Accuracy (\%)  & Accuracy (\%)\\
&  &  & Avg. class & Overall\\
\midrule
PointNet & $--$ & $3.47M$ & $86.13\pm{0.46}$ & $89.68 \pm{0.40}$ \\
All-SVD & $94.80$ & $0.18M$ & $67.27\pm{0.79}$  & $76.56\pm{1.28}$ \\
All-Adaptive Fastfood & $98.25$ & $0.06M$ & $--$ & $--$ \\
All-Butterfly & $93.32$  & $0.23M$ & $82.69\pm{0.42}$ & $\mathbf{88.17}\pm{0.09}$ \\
All-DeBut (Small) & $\mathbf{95.02}$ & $0.17M$ & $81.77\pm{1.78}$ & $87.85\pm{0.73}$ \\
All-DeBut (Large) & $93.50$ & $0.23M$ & $\mathbf{82.73}\pm{0.73}$ & $\mathbf{88.17}\pm{0.06}$ \\
\bottomrule
\end{tabular}
}
\end{table}

By substituting all layers with different structured linear transforms, the performance of various homogeneous models are listed in Table~\ref{tab:exp_1}. Two All-DeBut PointNets of different sizes are trained. The smaller one achieves an average class accuracy of $81.77\%$ and overall accuracy of $87.85\%$ while yielding the second-highest model-wise compression rate with only approximately $5\%$ parameters of the original model. Although the All-Butterfly counterpart has slightly higher accuracy ($\le1\%$), it has $35\%$ ($0.23M$ vs $0.17M$) more parameters. Besides, the standard Butterfly suffers the PoT and fixed input-output size constraints, resulting in expansion instead of compression in a few layers (i.e., CONV1 and input T-Net FC3). To maintain fairness, those layers are kept unchanged in our All-Butterfly implementation. To obtain a hardware-friendly homogeneous network, the implementation of standard Butterfly on some layers can lead to an increase in the number of parameters, whereas All-DeBut solution guarantees compression in every layer. 

By replacing the largest two layers of the smaller one with another set of DeBut chains, a bigger All-DeBut PointNet with a similar compression ($93.50\%$ vs $93.32\%$) as the All-Butterfly variant is built, which outperforms the All-Butterfly counterpart in average class accuracy ($82.73\%$ vs $82.69\%$) and ties in overall accuracy. Even though All-Adaptive Fastfood achieves the highest compression ratio of $98.25\%$, it is practically infeasible to train as the Fast Hadamard Transform is overly time-consuming. In our case, it takes approximately $1200$s to train only a single epoch on ModelNet40. 

\subsection{All-DeBut ResNet on CIFAR-100 and ImageNet}
\label{subsec:resnet_cifar100}

\begin{table}[t]
\caption{Results on CIFAR-100 of homogeneous networks obtained via different linear transforms, with ResNet110 backbone.}
\label{tab:exp_2}
\scriptsize
\centering
\setlength{\tabcolsep}{1.9mm}{
\begin{tabular}{lcccc}
\toprule
\multirow{2}{*}{Method} & \multirow{2}{*}{MC (\%)} & \multirow{2}{*}{\#Params} & Accuracy (\%)  & Accuracy (\%)\\
~ & ~ & ~ & Top-1 & Top-5\\
\midrule
ResNet110 & $--$ & $1.74M$ & $74.30 \pm{0.28}$ & $93.16 \pm{0.16}$ \\
All-SVD & $88.86$ & $0.19M$ & $35.46\pm{0.04}$ & $70.13\pm{0.79}$ \\
All-Butterfly & $91.31$ & $0.15M$ & $67.12\pm{0.21}$ & $90.29\pm{0.04}$ \\
All-DeBut (Small) & $\mathbf{91.89}$ & $0.14M$ & $68.07\pm{0.27}$ & $90.89\pm{0.16}$ \\
All-DeBut (Large) & $76.98$ & $0.40M$ & $\mathbf{71.64}\pm{0.17}$ & $\mathbf{92.29}\pm{0.34}$ \\
\bottomrule
\end{tabular}}
\end{table}

Having demonstrated the feasibility of All-DeBut on point cloud classification, we proceed to the \textbf{CIFAR-100} dataset comprising 100 classes, each with 500 training and 100 testing images. We use ResNet110 as the backbone. The results are in Table~\ref{tab:exp_2}. It can be observed that All-DeBut ResNet110 outperforms All-Butterfly ResNet in both top-1 ($68.07\%$ vs $67.12\%$) and top-5 ($90.89\%$ vs $90.29\%$) accuracies at the same compression ratio ($91.89\%$ vs $91.31\%$). ResNet with SVD replacement achieves a comparable model-wise compression but with much lower top-1 and top-5 accuracies of $35.46\%$ and $70.13\%$, respectively. Besides, All-DeBut ResNet demonstrates the flexibility of DeBut layers, in which one can increase the number of parameters by adapting different chains without giving up the homogeneity of All-DeBut networks. By allowing the model to hold more parameters ($0.40M$ vs $0.14M$) using a different set of chains, the All-DeBut ResNet can recover the top-1 accuracy to within $3\%$ and top-5 accuracy to even within $1\%$ versus the baseline.

An additional \textbf{ImageNet} experiment is conducted to demonstrate All-DeBut effectiveness on large-scale datasets. Specifically, we compared All-DeBut with the sophisticated Tucker-2 decomposition algorithm using a ResNet34 Backbone. The results, shown in Table~\ref{tab:imagenet_result}, indicate that under a similar compression ratio of $\sim70\%$, the All-DeBut network outperforms the Tucker-2 counterpart by an absolute gain of $>4\%$ in terms of Top-1 accuracy, making All-DeBut a viable choice for CNN compression.

\begin{table}[t]
\centering
\scriptsize
\caption{Results on ImageNet. ResNet34 is used as the backbone.}
\label{tab:imagenet_result}
\setlength{\tabcolsep}{2.5mm}{
\begin{tabular}{ccccc}
\toprule
Method    & MC (\%)        & \#Params (M) & Top-1 Acc (\%) & Top-5 Acc (\%) \\ 
\midrule
All-Debut & \textbf{71.89} & \textbf{6.13}         & \textbf{66.50} & \textbf{86.76} \\ 
Tucker-2  & 71.28          & 6.26         & 62.19          & 84.26          \\ 
\bottomrule
\end{tabular}
}
\end{table}

\subsection{DeBut Simulation on FPGA}
\label{subsec:fpga}


\begin{table}[t]
\caption{FPGA performance comparison using VGG16 as backbone. $W$\#$A$\# denotes the weight and activation bitwidths.}
\label{tab:fpga}
\scriptsize
\centering
\setlength{\tabcolsep}{7.3mm}{
\begin{tabular}{lcc}
\toprule
Method & Latency (ms) &  Bitwidth\\
\midrule
Baseline ({\tt conv2d}) & $23.39$  & $W8A8$ \\
All-DeBut& $\mathbf{13.62}$ & $W16A16$ \\
\bottomrule
\end{tabular}
}
\end{table}

To show the superiority of All-DeBut nets in hardware speedup, we simulate an All-DeBut VGG-16 on a Xilinx Ultra96-V2 FPGA using Vivado 2020.2 for High-Level Synthesis (HLS). Table~\ref{tab:fpga} compares our DeBut accelerator to baseline ({\tt conv2d}), running on the same hardware. The efficiency of DeBut stems from its structured sparsity, combined with a modular configurable network topology that can be scaled up or down subject to the desired accuracy. DeBut layers have emerged as a promising scheme to achieve $\approx 2\times$ speedup with much fewer parameters and operations over the traditional {\tt conv2d} layers.  




\section{Conclusion}
\label{sec:conclusion}
This paper has performed an in-depth study on the lately proposed Deformable Butterfly (DeBut) structured sparse matrix factorization. A newly devised automated DeBut chain generator largely obviates the labor in handcrafting DeBut chains and plays a crucial role in enabling the substitution of \emph{every} CNN or FC layer, for the first time, in a DNN to achieve a remarkable compression without much sacrifice in accuracy. Moreover, an intimate link is revealed between DeBut and a special form of depthwise separable convolution, thus explaining the effectiveness of DeBut. Experiments demonstrate the superiority of an All-DeBut network, added with FPGA benchmarking that showcases its hardware speedup and its promising candidacy for resource-limited edge deployment.%

{\flushleft{\textbf{Acknowledgement}}}

This work was supported in part by the Theme-based Research Scheme (TRS) project T45-701/22-R and in part by the General Research Fund (GRF) project 17209721 of the Research Grants Council (RGC), Hong Kong SAR.

\bibliographystyle{IEEEtran}
\bibliography{ref}

\begin{thebibliography}{10}
\providecommand{\url}[1]{#1}
\csname url@samestyle\endcsname
\providecommand{\newblock}{\relax}
\providecommand{\bibinfo}[2]{#2}
\providecommand{\BIBentrySTDinterwordspacing}{\spaceskip=0pt\relax}
\providecommand{\BIBentryALTinterwordstretchfactor}{4}
\providecommand{\BIBentryALTinterwordspacing}{\spaceskip=\fontdimen2\font plus
\BIBentryALTinterwordstretchfactor\fontdimen3\font minus \fontdimen4\font\relax}
\providecommand{\BIBforeignlanguage}[2]{{%
\expandafter\ifx\csname l@#1\endcsname\relax
\typeout{** WARNING: IEEEtran.bst: No hyphenation pattern has been}%
\typeout{** loaded for the language `#1'. Using the pattern for}%
\typeout{** the default language instead.}%
\else
\language=\csname l@#1\endcsname
\fi
#2}}
\providecommand{\BIBdecl}{\relax}
\BIBdecl

\bibitem{bracewell1986fourier}
R.~N. Bracewell and R.~N. Bracewell, \emph{The Fourier transform and its applications}.\hskip 1em plus 0.5em minus 0.4em\relax McGraw-Hill New York, 1986, vol. 31999.

\bibitem{phan2020stable}
A.-H. Phan, K.~Sobolev, K.~Sozykin, D.~Ermilov, J.~Gusak, P.~Tichavsk{\`y}, V.~Glukhov, I.~Oseledets, and A.~Cichocki, ``Stable low-rank tensor decomposition for compression of convolutional neural network,'' in \emph{European Conference on Computer Vision}.\hskip 1em plus 0.5em minus 0.4em\relax Springer, 2020, pp. 522--539.

\bibitem{yu2017compressing}
X.~Yu, T.~Liu, X.~Wang, and D.~Tao, ``On compressing deep models by low rank and sparse decomposition,'' in \emph{Proceedings of the IEEE Conference on Computer Vision and Pattern Recognition}, 2017, pp. 7370--7379.

\bibitem{wang2021kronecker}
D.~Wang, B.~Wu, G.~Zhao, M.~Yao, H.~Chen, L.~Deng, T.~Yan, and G.~Li, ``Kronecker cp decomposition with fast multiplication for compressing rnns,'' \emph{IEEE Transactions on Neural Networks and Learning Systems}, 2021.

\bibitem{frankle2018lottery}
J.~Frankle and M.~Carbin, ``The lottery ticket hypothesis: Finding sparse, trainable neural networks,'' \emph{arXiv preprint arXiv:1803.03635}, 2018.

\bibitem{frankle2020linear}
J.~Frankle, G.~K. Dziugaite, D.~Roy, and M.~Carbin, ``Linear mode connectivity and the lottery ticket hypothesis,'' in \emph{International Conference on Machine Learning}.\hskip 1em plus 0.5em minus 0.4em\relax PMLR, 2020, pp. 3259--3269.

\bibitem{wu2019compressing}
K.~Wu, Y.~Guo, and C.~Zhang, ``Compressing deep neural networks with sparse matrix factorization,'' \emph{IEEE transactions on neural networks and learning systems}, vol.~31, no.~10, pp. 3828--3838, 2019.

\bibitem{le2013fastfood}
Q.~Le, T.~Sarl{\'o}s, A.~Smola \emph{et~al.}, ``Fastfood-approximating kernel expansions in loglinear time,'' in \emph{Proceedings of the international conference on machine learning}, vol.~85, 2013.

\bibitem{yang2015deep}
Z.~Yang, M.~Moczulski, M.~Denil, N.~De~Freitas, A.~Smola, L.~Song, and Z.~Wang, ``Deep fried convnets,'' in \emph{Proceedings of the IEEE International Conference on Computer Vision}, 2015, pp. 1476--1483.

\bibitem{dao2019learning}
T.~Dao, A.~Gu, M.~Eichhorn, A.~Rudra, and C.~R{\'e}, ``Learning fast algorithms for linear transforms using butterfly factorizations,'' in \emph{International conference on machine learning}.\hskip 1em plus 0.5em minus 0.4em\relax PMLR, 2019, pp. 1517--1527.

\bibitem{dao2020kaleidoscope}
T.~Dao, N.~S. Sohoni, A.~Gu, M.~Eichhorn, A.~Blonder, M.~Leszczynski, A.~Rudra, and C.~R{\'e}, ``Kaleidoscope: An efficient, learnable representation for all structured linear maps,'' \emph{arXiv preprint arXiv:2012.14966}, 2020.

\bibitem{lin2021deformable}
R.~Lin, J.~Ran, K.~H. Chiu, G.~Chesi, and N.~Wong, ``Deformable butterfly: A highly structured and sparse linear transform,'' \emph{Advances in Neural Information Processing Systems}, vol.~34, 2021.

\bibitem{fukushima1982neocognitron}
K.~Fukushima and S.~Miyake, ``Neocognitron: A self-organizing neural network model for a mechanism of visual pattern recognition,'' in \emph{Competition and cooperation in neural nets}.\hskip 1em plus 0.5em minus 0.4em\relax Springer, 1982, pp. 267--285.

\bibitem{krizhevsky2012imagenet}
A.~Krizhevsky, I.~Sutskever, and G.~E. Hinton, ``Imagenet classification with deep convolutional neural networks,'' \emph{Advances in neural information processing systems}, vol.~25, pp. 1097--1105, 2012.

\bibitem{springenberg2014striving}
J.~T. Springenberg, A.~Dosovitskiy, T.~Brox, and M.~Riedmiller, ``Striving for simplicity: The all convolutional net,'' \emph{arXiv preprint arXiv:1412.6806}, 2014.

\bibitem{simonyan2014very}
K.~Simonyan and A.~Zisserman, ``Very deep convolutional networks for large-scale image recognition,'' \emph{arXiv preprint arXiv:1409.1556}, 2014.

\bibitem{he2016deep}
K.~He, X.~Zhang, S.~Ren, and J.~Sun, ``Deep residual learning for image recognition,'' in \emph{Proceedings of the IEEE conference on computer vision and pattern recognition}, 2016, pp. 770--778.

\bibitem{howard2017mobilenets}
A.~G. Howard, M.~Zhu, B.~Chen, D.~Kalenichenko, W.~Wang, T.~Weyand, M.~Andreetto, and H.~Adam, ``Mobilenets: Efficient convolutional neural networks for mobile vision applications,'' \emph{arXiv preprint arXiv:1704.04861}, 2017.

\bibitem{tan2019efficientnet}
M.~Tan and Q.~Le, ``Efficientnet: Rethinking model scaling for convolutional neural networks,'' in \emph{International Conference on Machine Learning}.\hskip 1em plus 0.5em minus 0.4em\relax PMLR, 2019, pp. 6105--6114.

\bibitem{tian2019crd}
Y.~Tian, D.~Krishnan, and P.~Isola, ``Contrastive representation distillation,'' \emph{arXiv preprint arXiv:1910.10699}, 2019.

\bibitem{wu2015modelnet}
Z.~Wu, S.~Song, A.~Khosla, F.~Yu, L.~Zhang, X.~Tang, and J.~Xiao, ``3d shapenets: A deep representation for volumetric shapes,'' in \emph{Proceedings of the IEEE conference on computer vision and pattern recognition}, 2015, pp. 1912--1920.

\bibitem{krizhevsky2009cifar100}
A.~Krizhevsky, G.~Hinton \emph{et~al.}, ``Learning multiple layers of features from tiny images,'' 2009.

\bibitem{deng2009imagenet}
J.~Deng, W.~Dong, R.~Socher, L.-J. Li, K.~Li, and L.~Fei-Fei, ``Imagenet: A large-scale hierarchical image database,'' in \emph{2009 IEEE conference on computer vision and pattern recognition}.\hskip 1em plus 0.5em minus 0.4em\relax Ieee, 2009, pp. 248--255.

\bibitem{qi2017pointnet}
C.~R. Qi, H.~Su, K.~Mo, and L.~J. Guibas, ``Pointnet: Deep learning on point sets for 3d classification and segmentation,'' in \emph{Proceedings of the IEEE conference on computer vision and pattern recognition}, 2017, pp. 652--660.

\end{thebibliography}

\end{document}